# Depth Map Estimation
# for Free-Viewpoint Television

Dawid Mieloch, Olgierd Stankiewicz and Marek Domański

*Abstract*—The paper presents a new method of depth estimation dedicated for free-viewpoint television (FTV). The estimation is performed for segments and thus their size can be used to control a trade-off between the quality of depth maps and the processing time of their estimation. The proposed algorithm can take as its input multiple arbitrarily positioned views which are simultaneously used to produce multiple inter-view consistent output depth maps. The presented depth estimation method uses novel parallelization and temporal consistency enhancement methods that significantly reduce the processing time of depth estimation. An experimental assessment of the proposals has been performed, based on the analysis of virtual view quality in FTV. The results show that the proposed method provides an improvement of the depth map quality over the state-of-the-art method, simultaneously reducing the complexity of depth estimation. The consistency of depth maps, which is crucial for the quality of the synthesized video and thus the quality of experience of navigating through a 3D scene, is also vastly improved.

## I. INTRODUCTION

IN free-viewpoint television (FTV) [41], on which we focus in this paper, a user can arbitrarily change her/his viewpoint at any time and is not limited to watch only the views acquired by cameras located around a scene. Views presented to the user are synthesized, i.e., rendered from a compact representation of a 3D scene [38]. One of the spatial representations of a 3D scene are depth maps [39], which are widely used not only in the context of free-viewpoint television systems [1], [29], [40], but also in 3D scene modeling [36], and machine vision applications [37], [51]. In FTV systems, the quality of depth maps is crucial for the quality of the synthesized video, and thus the quality of experience in the navigation through a 3D scene.

In general, depth can be captured with depth sensors or computed from stereoscopic correspondence if a given object is visible in more than one camera. Depth acquisition using depth sensors can be easily performed in real time [4]. However, using only one depth sensor can be insufficient for modeling the whole scene [4]. On the other hand, using more devices is difficult due to possible interferences between depth sensors [5] or from other infrared illumination sources. The presented problems, together with limited resolution and range of depth cameras, further limit the possible applications of depth sensors in FTV systems, although depth cameras and LIDARs underlie many improvements recently [59], [60]. Thus, the considerations of this paper are focused on depth estimation by multiview video analysis.

In FTV, the estimation of depth maps is not the final goal, but it is rather an important step in the process of production of virtual views. Therefore, through this paper, the quality of the depth maps is represented by the quality the virtual views synthesized using these depth maps. Such an approach is common in research on depth map estimation [62], [63] and was proposed also as a part of the 3D framework of the ISO/IEC MPEG group [61].

As this is discussed in Section II, the methods described in the references are not well-matched to the needs of FTV. Therefore, the goal of this paper is to present a new method of depth estimation designed for use in FTV**.** In order to provide very realistic viewing experience during virtual navigation, a new method of depth estimation has to meet a set of requirements that result from the characteristics of FTV.

FTV is characterized by a high number of cameras used for multiview video acquisition. Moreover, the resolution of cameras used in multiview systems constantly increases, especially for new virtual reality systems [30], [58]. At the same time, depth estimation is already one of the most complex parts of multiview video processing in free-viewpoint television systems, therefore, achieving higher quality comes at the cost of a further increase of complexity.

The characteristics of depth estimation for FTV require us not only to reduce the high complexity of the estimation, but also to ensure inter-view and temporal consistencies of depth maps. The virtual view synthesis uses depth maps and views from at least two nearest cameras [15], [38], [49]. The inter-view inconsistency of depth maps is related to independent depth estimation in neighboring views. Such independent estimation can cause inconsistency in the position of this object in a synthesized virtual view, which reduces both the objective and the subjective quality of the synthesized view [3]. The temporal consistency of depth maps, on the other hand, means that the values of depth in consecutive frames of depth maps change in accordance with the movement of objects in a scene, and what follows, the color and position of objects in a virtual view also change in accordance with their movement.

The variety of hitherto presented FTV systems [42] makes it

This work was supported by The National Centre for Research and Development, Poland under Project no. TANGO1/266710/NCBR/2015.

The authors are with the Chair of Multimedia Telecommunications and Microelectronics, Poznań University of Technology, 60965 Poznań, Poland (e-mail: dawid.mieloch@put.poznan.pl; olgierd.stankiewicz@put.poznan.pl; marek.domanski@put.poznan.pl).

difficult to develop a versatile depth map estimation method that could be successfully utilized in all such systems. FTV systems vary in the number and type of used cameras (from a few to hundreds), distances between them, and their positioning. Therefore, summarizing the presented characteristics of FTV systems, a new method of depth estimation has to be characterized by:
1) high quality of estimated depth maps, with particular emphasis on inter-view and temporal consistencies,
2) versatility of the estimation process, i.e. no assumptions about the number and positioning of cameras can be stated, and moreover, the method can be used for different scenes without any modifications,
3) processing time of the estimation that is shortened in comparison to the state-of-the-art methods that meet the abovementioned requirements (e.g., the new method has to provide the possibility of parallelization).

The novelty of the proposed method consists in addressing the above-mentioned characteristics by the joint application of many ideas, e.g., the use of image segmentation, depth estimation performed simultaneously for all views, cost function for improved inter-view consistency, enhancement of temporal consistency, and also the utilization of parallelization. The details of the proposed method are presented in Section III.

Although the idea of the proposed algorithm (i.e. inter-view consistent segment-based depth estimation) has already been briefly described in our previous work [32], here we present an extended framework in detail, together with much more comprehensive results, and we put particular emphasis on improvements that decrease the processing time of estimation. In order to maintain the consistency of description, the method will be presented in its entirety.

## II. STATE-OF-THE-ART DEPTH ESTIMATION METHODS

The simultaneous fulfillment of requirements concerning the inter-view and temporal consistencies of depth maps and, at the same time, achieving a relatively short processing time of estimation, is difficult without compromising the quality of the estimated depth [18]. For example, independent estimation of depth maps for each camera can be faster than the simultaneous estimation for all views [2], [26], however, the lowered number of views used during estimation causes the loss of inter-view consistency. Depth estimation can also be performed for input views with lowered resolution, nevertheless, the usage of low-resolution views decreases the accuracy of estimated depth maps and the resulting virtual view quality [33]. The loss of quality is especially visible near the edges or in highly textured regions [3], [19]. Even if an additional depth refinement by post-processing is used [34], [55], the quality of depth maps estimated in low resolution is still lower than for the estimation for the nominal resolution, also when virtual view synthesis methods designed for low resolution depth maps are used [35].

Depth estimation based on stereoscopic correspondence is very time-consuming, especially for global estimation methods that can provide depth maps of sufficient quality for view synthesis purposes (e.g., [7], [23]). The method described in [23], shown to be one of the fastest that uses global estimation, still requires about 10 minutes to estimate depth for a stereo pair (on a modern PC), and additionally requires input views to be rectified. Other methods also state this requirement [28], or are designed for multi-view systems of different characteristics than FTV systems, e.g. for light-field systems [21], [54], [56], or multi-camera arrays [50], which have much smaller distances between cameras. Inter-view and temporal consistencies are often also ensured, e.g. in [6] or [27], nevertheless, only for sequences acquired using a moving camera rig.

The use of depth estimation methods based on local estimation can ensure low complexity. Local estimation methods are very often suitable for real-time applications [18]. Although depth maps of relatively high quality [2], [26] or even depth maps that are inter-view and temporally consistent [24] can be estimated using such methods, the majority of these methods formulate additional requirements about the number or positioning of cameras. For example, methods [24] and [26] can be used only for a stereo pair, while [2] and [4] are strictly adjusted to multiview systems designed by their authors, reducing the usefulness of these methods in versatile free-viewpoint television systems.

Depth maps can also be estimated using an epipolar plane image [8], [31]. These methods enforce depth to be consistent in all views and are characterized by lower complexity than global estimation methods, but can be used only for dense multi-view systems.

More recently, a new interesting type of depth estimation methods was introduced, which uses convolutional neural networks to support the estimation process on the basis of previously prepared database of depth maps. Such data driven estimation, although it can represent the direction of future research in depth estimation, is still limited to specific applications (e.g., for soccer stadiums footage [22]), stereo pairs [16] or multi-view systems with a very narrow base [17], just like conventional methods presented earlier.

## III. PROPOSED GRAPH-BASED MULTIVIEW DEPTH ESTIMATION METHOD

### A. Overview of the Proposed Method

The novelty of the presented method of depth estimation, and its particular usefulness for free-viewpoint television systems, is a result of the joint application of the following ideas:
1) depth is estimated for segments instead for individual pixels, and thus the size of segments can be used to control the trade-off between the quality of depth maps and the processing time of estimation, without reducing the resolution of the estimated depth maps,
2) because of the utilization of the new formulation of the cost function, dedicated for segment-based estimation, estimation is performed for all views simultaneously and produces depths that are inter-view consistent with no assumptions about the positioning of views: any number of arbitrarily positioned cameras can be used,
3) although the segmentation is used, estimated depth is calculated on a per-pixel basis, because the

correspondence search is not limited to segment centers; the segmentation does not have to be consistent in all views, therefore, it is performed independently in each view, leading to the reduction of complexity,
4) in the proposed temporal consistency enhancement method, depth maps estimated in previous frames are utilized in the estimation of depth for the current frame, increasing the consistency of depth maps and simultaneously decreasing the processing time of estimation,
5) the proposed depth estimation framework uses a novel parallelization method that significantly reduces the processing time of graph-based depth estimation.

### B. Cost Function Formulation

The estimation of depth in the presented method is based on cost function minimization. The proposed cost function is a sum of two components, which are described in detail in Sections III-C and III-D: the intra-view discontinuity cost $V_{s,t}$ (a smoothing term) and the inter-view matching cost $M_{s,s'}$, responsible for the inter-view consistency of depth maps:

$$E(\underline{d}) = \sum_{c \in C}\{\sum_{c' \in D}\sum_{s \in S} M_{s,s'}(d_s) + \sum_{s \in S}\sum_{t \in T} V_{s,t}(d_s, d_t)\}, \quad (1)$$

where $\underline{d}$ is a vector of depth values for all segments in all views, C is a set of views for which depth is estimated, $c$ is a view used in the estimation, D is a set of views neighboring to view $c$, $c'$ is a view neighboring to view $c$, S is a set of segments of view $c$, $s$ is a segment in view $c$, $d_s$ is a currently considered depth of segment $s$, $s'$ is a segment in view $c'$, which corresponds to segment $s$ in view $c$ for depth $d_s$, $M_{s,s'}$ is an inter-view matching cost between segments $s$ and $s'$, T is a set of segments neighboring to segment $s$, $t$ is a segment neighboring to segment $s$, $V_{s,t}$ is an intra-view discontinuity cost between segments $s$ and $t$, $d_t$ is the currently considered depth of segment $t$.

Neighboring views $c'$ are two views: the nearest left view and the nearest right view of view $c$. For the leftmost and the rightmost view, the number of neighboring views used is limited to one: the nearest left or the nearest right view, depending which one is available.

In the considered scenario the optical axes of cameras do not have to be parallel. Therefore, in order to achieve inter-view consistency, the depth of a point has to be defined not as the distance from the plane of the camera that acquired this point, but as the distance from the plane of the center camera of the system [43] (for the sake of comprehension: the plane of a camera is a plane that contains the sensor of the camera).

A local minimum of the cost function (1) is estimated using the graph cut method [9] and the α-expansion method of the minimization for multi-label problems, described in details in [10].

Unlike in [9], where each node in the constructed graph represents one point of an input view, in our method a node corresponds to one segment. Nodes are connected by two types of links which correspond to the abovementioned intra-view discontinuity cost and the inter-view matching cost (Fig. 1).

The proposed segment-based estimation reduces the number of nodes in a graph in comparison with point-level estimation, making the process significantly faster. Simultaneously, depth maps in the presented method are still estimated in the same resolution as the nominal resolution of the input views, and because of the use of segments, the edges of objects in depth maps correspond to the edges of objects in input views. The number of segments, and therefore their size, is one of the estimation parameters and can be adjusted. The use of relatively small segments (i.e., of size of 20 samples or less) allows us to estimate high-quality depth significantly faster than in pixel-based estimation. On the other hand, the use of larger segments ensures an additional reduction of the processing time, at the expense of a minor loss of quality.

### C. Inter-view Matching Cost

In order to achieve the inter-view consistency of estimated depth maps, the matching cost is not calculated independently for each single view. Instead, the conventional matching cost is replaced with the inter-view matching cost $M_{s,s'}(d_s)$, which is defined between a pair of segments $s$ and $s'$ that correspond to one another for the currently considered depth $d_s$ (presented as the dotted arrow in Fig. 1).

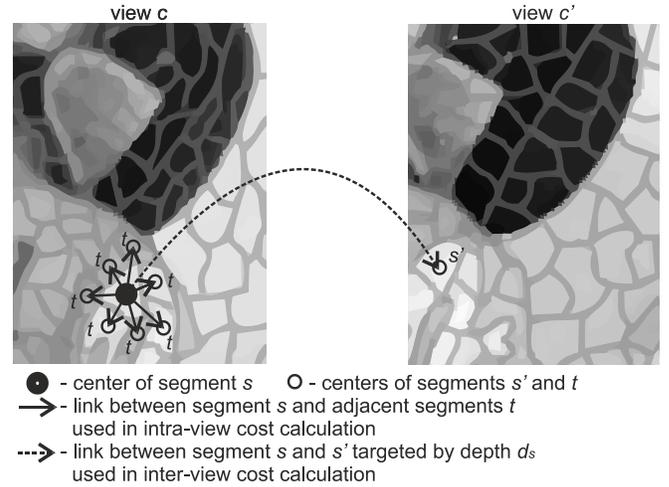

● - center of segment $s$   ○ - centers of segments $s'$ and $t$
→ - link between segment $s$ and adjacent segments $t$ used in intra-view cost calculation
--▶ - link between segment $s$ and $s'$ targeted by depth $d_s$ used in inter-view cost calculation

Fig. 1.  Visualization of intra-view discontinuity cost and inter-view matching cost for an exemplary segment $s$ for depth estimation performed for 2 views.

The proper matching of whole segments from different views is a difficult operation. Moreover, for the presented method no assumptions about the positioning of views are made. Therefore, the segmentation of the same object in neighboring views may significantly vary and result in different shapes and sizes of the corresponding segments. These differences are especially big when the optical axes of cameras are not parallel, because corresponding parts of a scene can be visible from different angles in neighboring cameras. Inter-view consistent segmentation would require correct depth maps, obviously not available at the beginning of depth estimation.

In order to avoid the abovementioned difficulties, the inter-view matching cost is calculated in the pixel-domain in a small window around the center of a segment and the corresponding point in a neighboring view. The core of the inter-view matching cost, denoted as $m_{s,s'}(d_s)$, is:

$$m_{s,s'}(d_s) = \frac{1}{size(W)} \sum_{w \in W} \left\| [YC_bC_r]_{\mu_s+w} - [YC_bC_r]_{T[\mu_s]+w} \right\|_1, \quad (4)$$

where W is a set of points in the window of the size specified by the user, $w$ is a point in window W, $\|\cdot\|_1$ denotes L1 distance, $\mu_s$ is the center of segment $s$, $[YC_bC_r]_{\mu_s+w}$ is the vector of Y, C$_b$, C$_r$ color components of center $\mu_s$ of segment $s$, $T[\cdot]$ is a 3D transform obtained from intrinsic and extrinsic parameters of cameras, $[YC_bC_r]_{T[\mu_s]+w}$ is the vector of Y, C$_b$, C$_r$ color components of the point in view $c'$ corresponding to center $\mu_s$ of segment $s$ in view $c$.

In order to achieve inter-view consistent depth maps, the value of the inter-view matching cost $M_{p,p'}(d_p)$ is calculated as:

$$M_{s,s'}(d_s) = \begin{cases} \min\{0, m_{s,s'}(d_s) - K\} & \text{if } d_s = d_{s'} \\ 0 & \text{if } d_s \neq d_{s'} \end{cases}, \quad (5)$$

where $s$ is a segment in view $c$, $d_s$ is the currently considered depth of segment $s$, $s'$ is a segment in view $c'$, which corresponds to segment $s$ in view $c$ for the currently considered depth $d_s$, $d_{s'}$ is the currently considered depth of segment $s'$. $K$ must be a positive constant [44]. In the presented method we use $K = 30$, as it provided the high quality of estimated depth maps in the preliminary tests for all tested sequences.

The presented definition of the inter-view matching cost does not require segmentation that is inter-view consistent in neighboring views. The center of a segment can correspond in the neighboring view to any point, not necessarily the center of a segment. Therefore, the presented pixel-domain matching lets us estimate the depth with high precision, simultaneously reducing the processing time of estimation, as the matching is not performed for all points, but only for centers of segments.

### D. Intra-view Discontinuity Cost

The intra-view discontinuity cost is calculated between all adjacent segments within a view (presented as the black solid arrows in Fig. 1). The cost is calculated as follows:

$$V_{s,t}(d_s, d_t) = \beta \cdot |d_s - d_t|, \quad (2)$$

where $\beta$ is a smoothing coefficient, $d_s$ and $d_t$ are the currently considered depths of adjacent segments $s$ and $t$. The smoothing coefficient $\beta$ is calculated adaptively using $\beta_0$, which is the initial smoothing coefficient provided by the user, and the similarity of segments $s$ and $t$ – the L1 distance (depicted as $\|\cdot\|_1$) between vectors $[\hat{Y}\hat{C}_b\hat{C}_r]_s$ and $[\hat{Y}\hat{C}_b\hat{C}_r]_t$ of average Y, C$_b$ and C$_r$ color components of the abovementioned segments:

$$\beta = \beta_0 / \left\| [\hat{Y}\hat{C}_b\hat{C}_r]_s - [\hat{Y}\hat{C}_b\hat{C}_r]_t \right\|_1. \quad (3)$$

When the similarity of adjacent segments is low, the smoothing coefficient also drops in value and thus the depths of these segments are not penalized for being discontinuous.

### E. Temporal Consistency Enhancement

In natural video sequences, only a small part of an acquired scene considerably changes in consecutive frames, especially when cameras are not moving during the acquisition of video. The idea of the proposed temporal consistency enhancement of depth estimation is to calculate a new value of depth only for the segments that changed (in terms of their color) in comparison with the previous frame.

The proposed temporal consistency enhancement method allows us to mark segments as unchanged in consecutive frames. These segments are used in the calculation of the intra-view discontinuity and the inter-view matching cost for other segments, but are not represented by any node in the structure of the optimized graph. It reduces the number of nodes in the graph, making the optimization process significantly faster and, on the other hand, increases the temporal consistency of estimated depth maps.

In the first frame of a depth map, denoted as an "I-type" depth frame (by analogy to video compression terminology), the estimation is performed for all segments, as described in the previous sections. The following frames ("P-type" depth frames) can utilize depth information from the preceding P-type depth frame and the I-type depth frame.

Segment $p$ is marked as unchanged in two cases: if all components of vector $[\hat{Y}\hat{C}_b\hat{C}_r]_s$ changed less than the set threshold $T_b$ in comparison with segment $s_B$, which is collocated segment in the previous P frame, or, if all components of the abovementioned vector changed less than threshold $T_I$ in comparison with segment $s_I$ – a collocated segment in the I frame. If any of these two conditions are met, then segment $s$ adopts the depth from the segment $s$ or $s_I$ (depending on which condition was fulfilled). Thresholds were estimated in preliminary tests and set as $T_P = 3$ and $T_I = 1$.

The introduction of two reference depth frames has a beneficial impact on the visual quality of virtual navigation in free-viewpoint television. Firstly, the adoption of depth from the previous P-type depth frame allows us to use the depth of objects that changed their position over time. On the other hand, the adoption of depth from the I-type depth frame minimizes the flickering of depth in the background.

## IV. DEPTH ESTIMATION FRAMEWORK IMPLEMENTATION DETAILS

In this section, we present the methods and solutions used in the realized implementation of the proposed depth estimation.

### A. Parallelization Method

In order to decrease the overall processing time of depth estimation in the presented method, the estimation is performed in parallel. In our proposal, each of $n$ threads estimates a depth map with an $n$-times lower number of depth levels (depth levels are planes that are parallel to the plane of the camera). In the presented method, depth levels can be distributed onto threads in two ways: depth levels can be interleaved or divided into blocks (Fig. 2).

The distribution of depth levels has an influence on the processing time and quality of the estimated depth maps. If objects of an acquired scene are placed more densely in some ranges of depths, the estimation for corresponding depth levels is longer. Therefore, if the depth levels are divided into blocks, the estimation for some threads can be longer, increasing the overall processing time of depth estimation. On the other hand, when depth levels are interleaved, the processing time of estimation for all threads is nearly equal, but the estimated depths tend to be less smooth. The dependency between the type of parallelization and the performance of depth estimation method was tested in one of the performed experiments presented in Section V.

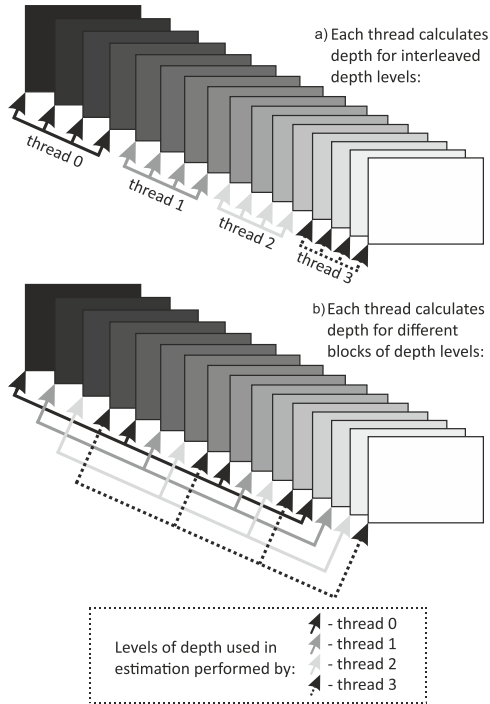

Fig. 2. Two examples of different depth level distributions over threads in the proposed method: a) depth levels are divided into blocks, b) depth levels are interleaved. Each rectangle represents a different level of the depth of a scene.

Depth maps with a reduced number of depth levels that were calculated by different threads have to be merged into one depth map. The merging process is performed in a similar way as depth estimation [using the cost function (1)], but only two levels of depth are considered for each segment – i.e., the depth of a segment from thread $t$ or the depth from thread $t+1$ (Fig. 3). Only two depth maps can be merged into one by one thread during the merging cycle. Therefore, for $n$ threads, $\lceil log2(n) \rceil$ of additional cycles are needed to estimate the final depth map with all depth levels.

Of course, even without the use of parallelization, all cores of the CPU can also be used for depth estimation, e.g., each core can perform the estimation of depth for different sets of input views (e.g., for each 5 cameras of the system), or for different frames of the sequence. Unfortunately, when many standalone depth estimation processes are performed, it results in the loss of inter-view consistency, or temporal consistency of estimated depth maps. When the proposed parallelization is used, both inter-view and temporal consistency of depth maps, which are fundamental for the quality of virtual view synthesis, are preserved.

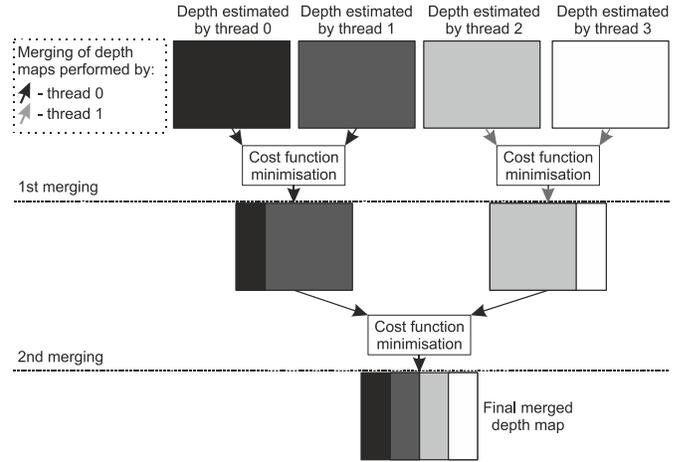

Fig. 3. Depth map merging process for the case of 4-thread parallelization.

### B. Optimization Method

The proposed method utilizes the graph cuts method to estimate depth maps [9], [10]. As it was proven in [25], the improvement of problem formulation has a significantly larger influence on depth estimation performance than the selection of the optimization method. Additionally, the graph cuts method, in comparison with belief propagation, the competitive method of global optimization, handles the penalties between nodes of the graph in a better way [25]. Therefore, in the proposed method of depth estimation, where graph construction is strictly based on dependencies between segments, the use of the graph cuts method is advisable and favorable.

### C. Segmentation

The proposed method of depth estimation can be used with any superpixel segmentation method. The authors decided to use the SNIC method (Simple Non-Iterative Clustering [20]), because the properties of SNIC meet the characteristics of the proposed depth estimation method: segments represent small regions, not the whole objects, and the number of segments can be freely changed. The SNIC method has also been shown to have low complexity (which reduces the overall processing time of depth estimation) and achieve one of the lowest segmentation errors when compared to state-of-the-art methods, which positively influences the representation of edges of objects in depth maps.

In the presented framework of depth estimation, instead of the $CIELAB$ space, in order to avoid the recalculation of color space, the segments are calculated using the $YC_bC_r$ color space. The used parameters of the segmentation are the compactness factor $m = 5$, and 8-connected segments.

## V. EXPERIMENTAL RESULTS

### A. Assessment of the Quality of Depth Maps

The quality of depth maps is measured indirectly, through the virtual view synthesis. The available databases with ground-truth depth maps do not correspond to the characteristics of free-viewpoint television. The newest Middlebury database [52] is widely used by the research community and allows us to easily evaluate the performance of a depth estimation method and compare it with other methods. Unfortunately, the comparison of depth estimation methods in this database is performed for a set of rectified stereo-pair images acquired using two cameras with parallel optical axes, while in free viewpoint television systems any number of arbitrarily positioned cameras can be used. Moreover, the dataset includes only one frame for each scene, therefore, the temporal consistency of depth maps, which is a significant part of research presented in this paper, also cannot be measured using this database.

Other databases of ground-truth depth maps (e.g., one of the newest databases – the ETH3D Benchmark [53]) also focus on the use of multi-camera systems of different properties than FTV, e.g., on moving camera rigs, or on a 3D reconstruction of static scenes.

However, for an end user, the quality of virtual views expresses the overall quality of a free-viewpoint television system. Therefore, virtual views are a good determinant of the performance of a depth estimation method.

In the experiments, a set of 8 multiview test sequences of varied character and arrangement of cameras are used. Sequences, their resolutions, views used in experiments and their sources are presented in Table I.

TABLE I
TEST SEQUENCES USED IN EXPERIMENTS

| Test sequence | Resolution | Used views | Sequence source |
|---|---|---|---|
| Ballet Breakdancers | 1024×768 | 0 to 7 | Microsoft Research [11] |
| BBB Butterfly BBB Rabbit | 1280×768 | 6, 12, 19, 26, 32, 38, 45, 52 | Holografika [12] |
| Poznań Blocks Poznań Blocks2 Poznań Fencing2 Poznań Service2 | 1920×1080 | 0 to 7 | Poznań University of Technology [13][14] |

In the conducted experiments not only we compare our method with the state-of-the-art graph-based depth estimation method DERS [7] (Section V-B1), but we also determine the performance of the presented method for different numbers of segments (Section V-B2), and for different numbers of views used in the estimation (Section V-B3). The performance of the presented parallelization methods and temporal consistency enhancement is also tested (Sections V-B4 and V-B5 respectively).

The scheme of measuring the depth map quality is presented in Fig. 4. The synthesis of a virtual view placed in the position of the acquired view 2 is performed using neighboring views 1 and 3 and corresponding estimated depth maps. The synthesized virtual view is compared with the acquired view 2 and PSNR of luminance is calculated and averaged for 50 frames for each test sequence. In the experiments, besides the quality of estimated depth maps, we also measure the processing time of estimating depth per one frame and view of a sequence. There are 5 views used during estimation, except for the analysis of the influence of the number of views on the quality of virtual views (Section V-B3). In order to decrease the overall processing time of the estimation, temporal consistency enhancement is turned on in all experiments (the number of P-type depth frames between I-type depth frames is 9), except for comparison with DERS, where the enhancement is not used. It is worth noting that in the case of free navigation, the virtual views are estimated from two or more nearest views, e.g., the virtual view between acquired views 1 and 2 is usually synthesized using exactly these two views. The nearest acquired view is, in the worst possible case, distant from the virtual view by half of the distance between cameras. The distance between the position of the virtual view and the acquired views has a significant impact on the quality of virtual views [45]. Here, the distance between views used for view synthesis is larger, therefore the overall quality of virtual navigation obtained from estimated depth maps would be noticeably higher than presented in the experiments.

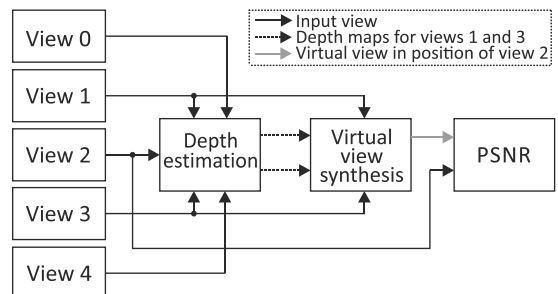

Fig. 4. The scheme of calculating PSNR of the virtual view synthesized using depth maps estimated in the experiment.

All experiments were performed on one thread of Intel Core i7-5820K CPU (3.3 GHz clock) machines equipped with 64 GB of operational memory (except for the test of the parallelization method, where the number of used cores varied from 1 to 6). The size of a block in the inter-view matching cost is 3×3 and the estimation is performed for 250 levels of depth. The synthesis of virtual views is performed using the View Synthesis Reference Software developed by the MPEG community [15].

*B. Results*

*1) Comparison with DERS*

The presented method is compared with the state-of-the-art Depth Estimation Reference Software developed by the MPEG community [7]. DERS is the graph-based method available for the researchers in its entirety and it states no assumptions about the positioning of cameras. Therefore, DERS is a reasonable reference depth estimation method for the presented framework.

For HD test sequences the number of segments used in the proposed method is 100,000, while for sequences with the lower resolution, in order to ensure a similar size of segments for all sequences, the number of segments is 50,000. The temporal consistency enhancement is turned off and the estimation is performed without parallelization. Other parameters of estimation are the same for both methods.

Table II presents the results of the experiment. For all tested sequences the quality of virtual views synthesized using depth maps estimated with the proposed method is higher than for depth maps from DERS, with the maximum gain in the quality equal to more than 5 dB. The average gain for all sequences is 2.63 dB. The lowest PSNR of a virtual view for the DERS is below 22 dB, while for the proposed method the lowest PSNR is 25.5 dB. For the proposed method only for one sequence the PSNR is below 27 dB. For DERS there are five such sequences. The visual comparison of depth maps for the proposed method and DERS, together with synthesized virtual views, is available in the video attached to this paper as supplementary material.

TABLE II
COMPARISON OF QUALITY OF VIRTUAL VIEWS SYNTHESIZED USING DEPTH MAPS ESTIMATED USING THE PROPOSED METHOD AND THE REFERENCE METHOD DERS [7]

| Test sequence | PSNR of virtual view [dB] | | | Processing time of depth estimation per one view [s] | | |
|---|---|---|---|---|---|---|
| | DERS | Proposal | Gain | DERS | Proposal | Ratio |
| Ballet | 27.93 | 28.69 | 0.76 | 882 | 499 | 57% |
| Breakdancers | 31.13 | 32.19 | 1.06 | 949 | 255 | 27% |
| BBB Butterfly | 29.97 | 33.20 | 3.23 | 593 | 279 | 47% |
| BBB Rabbit | 22.59 | 27.21 | 4.62 | 744 | 92 | 12% |
| Poznań Blocks | 21.97 | 27.20 | 5.23 | 1445 | 313 | 22% |
| Poznań Blocks2 | 25.67 | 28.12 | 2.45 | 1060 | 210 | 20% |
| Poznań Fencing2 | 26.74 | 28.60 | 1.86 | 2254 | 391 | 17% |
| Poznań Service2 | 23.69 | 25.51 | 1.82 | 2780 | 305 | 11% |
| | | *Average:* | 2.63 | | *Average:* | 27% |

Furthermore, the estimation process is, on average, almost 4 times faster for the presented method. What is important, the reduction of the processing time of estimation is highest for the HD sequences, therefore, the proposed method can be effectively used with high resolution cameras. It is the effect of the use of segmentation in depth estimation – the complexity of estimation in the proposed method is dependent on the number of segments, not on the resolution of input views.

What should be stressed again, the presented times of depth estimation are obtained for the core version of the presented method and do not take into account enhancements and methods (tested in the further part of this section) that significantly reduce the processing time of the estimation.

*2) Results for Different Numbers of Segments*

In the next experiment, the influence of the number of segments used in depth estimation on the quality of a virtual view is tested. The number of segments varied from 1,000 to 150,000.

The results of the experiment, averaged for all sequences, are presented in Fig. 5. As it can be clearly observed, the more segments are used in the estimation, the higher quality of depth maps can be achieved. However, the use of more than 100,000 insignificantly increases the quality of depth maps, at the cost of a considerable increase of the estimation time. The results for individual sequences are presented in Table IV in the Appendix. The visual comparison of depth maps estimated for different numbers of segments, together with synthesized virtual views, is also available in the video attached to this paper as supplementary material.

When only 1,000 segments are used, the quality of depth maps is equal to the average quality of depth maps estimated using the DERS method, but the time needed for the estimation process is significantly shorter and equal to only 2 seconds.

The highest increase of the quality of depth maps can be seen between estimations performed for 1,000 and 25,000 segments per view. Despite the number of segments increasing 25-fold, the average processing time of estimation increases only six-fold. On the other hand, increasing the number of segments above 100,000 does not change the quality of depth maps significantly (only 0.1 dB), but the mean processing time of estimation is noticeably longer.

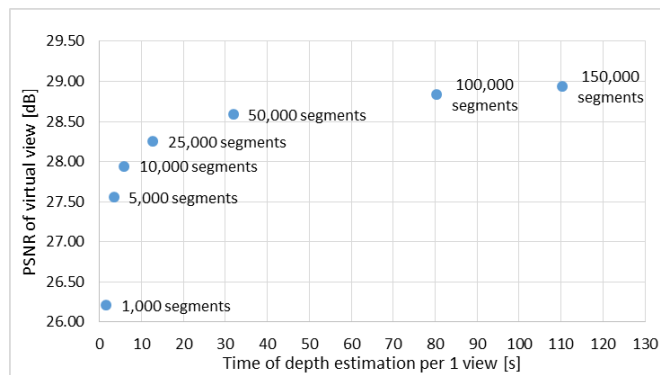

Fig. 5. The average quality of a virtual view synthesized using depth maps estimated for different numbers of segments per one view and processing times of depth estimation.

*3) Results for Different Numbers of Views*

The influence of the number of views used in the estimation process on the quality of the estimated depth maps is also tested. The number of views varies from 3 to 8 and is limited by the number of views available in test sequences.

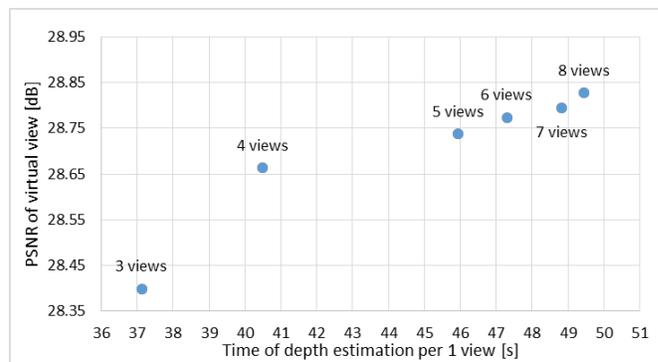

Fig. 6. The average quality of a virtual view synthesized using depth maps estimated for different numbers of views used in the estimation process and processing times of depth estimation.

The results presented in Fig. 6 show that the use of more than 5 views changes the measured quality of virtual views and the processing time of estimation only to a small extent. However, the use of all available views increases the inter-view consistency of estimated depth maps, therefore, we recommended performing the estimation for all views simultaneously to ensure the high quality of free navigation.

The results for individual sequences are presented in Table V in the Appendix.

*4) Results for Different Types of Parallelization*

The presented parallelization method is tested in two variants: blocks of depth levels (Fig. 2a) and interleaved levels of depth (Fig. 2b). The number of used threads varies from 1 to 6, and is limited by the number of standalone cores in the used CPU. The results of the experiment (Fig. 7) confirm that if the levels of depth are distributed onto threads as blocks of depth levels, the processing time of the estimation is slightly longer than for interleaved levels of depth, but the difference in the quality increases with the number of threads used.

Even when 6 threads are used, the quality decrease in comparison with the estimation without parallelization is insignificant (around 0.1 dB) but the processing time of the estimation decreases 4.5-fold. The results for individual sequences are presented in Table VI in the Appendix. The visual comparison of depth maps estimated using proposed parallelization method, together with synthesized virtual views, is also available in the video attached to this paper as supplementary material.

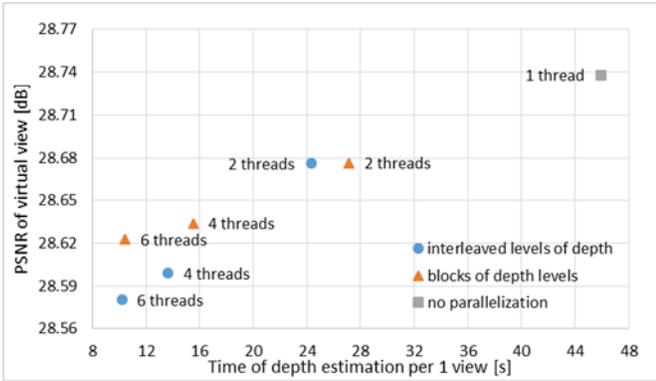

Fig. 7. The average quality of a virtual view synthesized using depth maps estimated for different parallelization cases and processing times of depth estimation.

Moreover, both the inter-view and temporal consistency of depth maps, which are fundamental for the quality of virtual view synthesis, are preserved when the proposed parallelization is used. The method is fully scalable, so the constantly increasing number of cores in modern CPUs can be fully utilized for further reduction of the processing time of depth estimation.

*5) Results for Different Numbers of B Type Depth Frames*

Here, we present the performance of the proposed temporal consistency enhancement of the proposed depth estimation method. The number of frames is 50, as in all conducted experiments, and the number of used P-type depth frames between I-type frames varies from 0 to 49.

Fig. 8 shows the results of the performed experiment. The temporal consistency enhancement significantly reduces the processing time of estimation (10 times when only one I frame is used) with a negligible decrease of the objective quality (less than 0.3 dB). The results for individual sequences are presented in Table VII in the Appendix. The visual comparison of depth maps estimated using proposed temporal consistency enhancement method, together with synthesized virtual views, is also available in the video attached to this paper as supplementary material.

The results presented above relate only to the quality of virtual views and do not express the increase of temporal consistency of depth maps. As it was presented earlier [46], the size of depth maps after encoding is one of the objective measures of their temporal consistency. In this article, we focus on the quality of free navigation for a user of the FTV system, therefore, in order to measure the increase of the temporal consistency of depth maps, synthesized virtual views are compressed with the HEVC encoder. The lack of temporal consistency of depth maps results in visible flickering of the virtual view. Therefore, the lower the temporal consistency of depth maps, the lower the efficiency of encoding of virtual views.

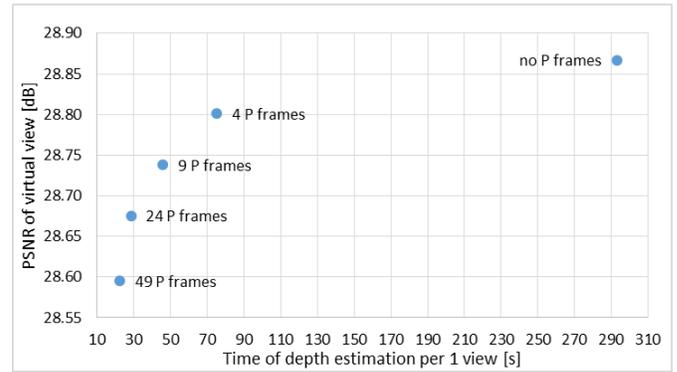

Fig. 8. The average quality of a virtual view synthesized using depth maps estimated for different numbers of P-type depth frames between I-type depth frames and processing times of depth estimation.

The encoder is set in the low-delay mode, so only the first frame of virtual views is encoded as an intra frame. Such settings of the encoder increase the influence of temporal consistency of the encoded sequence on the final bitrate. In the experiments, we use the HM 16.15 framework [47] using MPEG common test conditions (with the exception of used test sequences) and software reference configurations.

Table III presents the results of encoding virtual views synthesized using depth maps with different numbers of P-type depth frames. The results are expressed as average luma bitrate reductions calculated using the Bjøntegaard [48] metric in comparison to a virtual view synthesized with depth that was not temporally enhanced. The detailed results for all QPs that include a bitrate and PSNR after encoding are presented in Table VIII in the Appendix.

For all tested sequences, the use of the proposed temporal consistency enhancement of depth maps results in bitrate reduction for all encoded virtual views. The average reduction is even larger than 30% when the number of P-type depth frames is equal to 49 (therefore only one I-type depth frame is used in the whole sequence). It indicates that the proposed technique of temporal consistency enhancement significantly increases the temporal consistency of depth maps, because the performance of the encoder in low-delay mode is vastly dependent on temporal prediction. The results also show another advantage of temporal consistency of depth maps in FTV system – the reduction of the bitrate required to send a virtual viewpoint to an end user.

TABLE III
AVERAGE LUMA BITRATE REDUCTIONS OF ENCODED VIRTUAL VIEWS SYNTHESIZED USING DEPTH MAPS ESTIMATED FOR DIFFERENT NUMBERS OF P-TYPE DEPTH FRAMES BETWEEN I-TYPE DEPTH FRAMES

| Test sequence | Number of P type depth frames between I-type depth frames | | | |
|---|---|---|---|---|
| | 4 | 9 | 24 | 49 |
| | Encoded virtual views bitrate reduction compared to virtual views synthesized using depth maps with no P-type depth frames | | | |
| Ballet | -14.4% | -19.0% | -20.5% | -20.8% |
| Breakdancers | -24.9% | -33.1% | -34.4% | -34.1% |
| BBB Butterfly | -18.8% | -29.2% | -34.6% | -38.3% |
| BBB Rabbit | -7.9% | -8.0% | -11.3% | -19.7% |
| Poznań Blocks | -4.5% | -5.7% | -7.3% | -6.1% |
| Poznań Blocks2 | -30.7% | -35.0% | -36.9% | -39.9% |
| Poznań Fencing2 | -30.7% | -53.5% | -68.8% | -75.8% |
| Poznań Service2 | -21.6% | -30.0% | -23.5% | -21.7% |
| *Average*: | -19.2% | -26.7% | -29.7% | -32.0% |

## VI. CONCLUSIONS

In this paper a new method of depth estimation designed for free-viewpoint television systems was presented. The estimation is based on the minimization of a new cost function formulated over segments instead of over individual pixels. The use of segmentation reduces the overall complexity of depth estimation and, at the same time, to estimate depth maps in their nominal resolution.

The presented method allows us to estimate depth maps of high quality that are inter-view and temporally consistent, which ensures the high quality of experience in free navigation. Moreover, no assumptions about the number and positioning of cameras used for the estimation are stated, making the method useful for any FTV system.

The performed comprehensive tests of the method show that the presented enhancements of the depth estimation method significantly reduce the processing time of the estimation in comparison to the reference state-of-the-art method. Depth maps which can be successfully used for virtual view synthesis purposes can be estimated in a time shorter than 5 seconds per one frame. Therefore, the overall time from the acquisition of an FTV sequence to the presentation of free navigation becomes acceptable, which can bring FTV systems closer to practical use. The particular usefulness of the presented depth estimation method was already confirmed by its implementation in an operational FTV system developed by Chair of Multimedia Telecommunications and Microelectronics of Poznań University of Technology [57].

## APPENDIX

TABLE IV
THE QUALITY OF VIRTUAL VIEWS SYNTHESIZED USING DEPTH MAPS ESTIMATED FOR DIFFERENT NUMBERS OF SEGMENTS

| Test sequence | Number of segments | | | | | | |
|---|---|---|---|---|---|---|---|
| | 1,000 | 5,000 | 10,000 | 25,000 | 50,000 | 100,000 | 150,000 |
| | Mean PSNR of the virtual views [dB] | | | | | | |
| Ballet | 26.55 | 28.02 | 28.30 | 28.53 | 28.64 | 28.83 | 28.85 |
| Breakdancers | 29.12 | 30.85 | 31.40 | 31.67 | 32.00 | 32.14 | 32.15 |
| BBB Butterfly | 30.26 | 32.07 | 32.36 | 32.79 | 33.08 | 33.23 | 33.25 |
| BBB Rabbit | 24.84 | 26.08 | 26.64 | 26.91 | 27.14 | 27.43 | 27.73 |
| Poznań Blocks | 22.43 | 24.55 | 25.27 | 25.74 | 26.60 | 27.14 | 27.26 |
| Poznań Blocks2 | 25.94 | 26.93 | 27.16 | 27.54 | 27.87 | 28.10 | 28.19 |
| Poznań Fencing2 | 26.17 | 27.29 | 27.60 | 27.92 | 28.19 | 28.43 | 28.61 |
| Poznań Service2 | 24.35 | 24.66 | 24.80 | 24.93 | 25.17 | 25.38 | 25.51 |
| *Average*: | 26.21 | 27.55 | 27.94 | 28.25 | 28.59 | 28.84 | 28.94 |

TABLE V
THE QUALITY OF VIRTUAL VIEWS SYNTHESIZED USING DEPTH MAPS ESTIMATED FOR DIFFERENT NUMBERS OF VIEWS USED IN ESTIMATION

| Test sequence | Number of views | | | | | |
|---|---|---|---|---|---|---|
| | 3 | 4 | 5 | 6 | 7 | 8 |
| | Mean PSNR of the virtual views [dB] | | | | | |
| Ballet | 28.68 | 28.78 | 28.63 | 28.93 | 28.98 | 29.01 |
| Breakdancers | 31.98 | 32.05 | 31.99 | 32.24 | 32.14 | 32.28 |
| BBB Butterfly | 32.32 | 33.13 | 33.08 | 33.63 | 33.62 | 33.55 |
| BBB Rabbit | 26.82 | 27.25 | 27.13 | 27.60 | 27.27 | 27.44 |
| Poznań Blocks | 25.95 | 26.37 | 27.13 | 26.32 | 26.66 | 26.66 |
| Poznań Blocks2 | 28.14 | 28.27 | 28.09 | 27.73 | 27.94 | 27.89 |
| Poznań Fencing2 | 28.12 | 28.15 | 28.42 | 28.64 | 28.68 | 28.72 |
| Poznań Service2 | 25.14 | 25.27 | 25.37 | 25.06 | 25.03 | 25.02 |
| *Average*: | 28.39 | 28.66 | 28.73 | 28.77 | 28.79 | 28.82 |

TABLE VI
THE QUALITY OF VIRTUAL VIEWS SYNTHESIZED USING DEPTH MAPS ESTIMATED FOR DIFFERENT PARALLELIZATION TYPES

| Test sequence | Parallelization type | | | | | |
|---|---|---|---|---|---|---|
| | None | Interleaved levels of depth | | Blocks of depth levels | | |
| | Number of threads used in depth estimation | | | | | |
| | 1 | 2 | 4 | 6 | 2 | 4 | 6 |
| | Mean PSNR of the virtual views [dB] | | | | | |
| Ballet | 28.64 | 28.64 | 28.72 | 28.71 | 28.34 | 28.30 | 28.18 |
| Breakdancers | 32.00 | 32.05 | 32.06 | 31.98 | 31.93 | 31.87 | 31.87 |
| BBB Butterfly | 33.08 | 33.09 | 32.95 | 32.85 | 33.20 | 33.13 | 33.19 |
| BBB Rabbit | 27.14 | 27.11 | 27.10 | 27.07 | 27.04 | 26.97 | 27.02 |
| Poznań Blocks | 27.14 | 26.67 | 26.33 | 26.47 | 27.12 | 27.08 | 27.04 |
| Poznań Blocks2 | 28.10 | 28.07 | 28.01 | 27.97 | 28.10 | 28.09 | 28.10 |
| Poznań Fencing2 | 28.43 | 28.38 | 28.22 | 28.22 | 28.41 | 28.36 | 28.36 |
| Poznań Service2 | 25.38 | 25.39 | 25.40 | 25.36 | 25.27 | 25.27 | 25.22 |
| *Average*: | 28.74 | 28.68 | 28.60 | 28.58 | 28.68 | 28.63 | 28.62 |

TABLE VII
THE QUALITY OF VIRTUAL VIEWS SYNTHESIZED USING DEPTH MAPS ESTIMATED FOR DIFFERENT NUMBERS OF P-TYPE DEPTH FRAMES BETWEEN I-TYPE DEPTH FRAMES

| Test sequence | Number of P-type depth frames between I type depth frames | | | | |
|---|---|---|---|---|---|
| | 0 | 4 | 9 | 24 | 49 |
| | Mean PSNR of the virtual views [dB] | | | | |
| Ballet | 28.69 | 28.68 | 28.63 | 28.74 | 28.75 |
| Breakdancers | 32.19 | 32.13 | 31.99 | 31.95 | 31.75 |
| BBB Butterfly | 33.20 | 33.14 | 33.08 | 32.97 | 32.93 |
| BBB Rabbit | 27.21 | 27.16 | 27.13 | 27.12 | 27.08 |
| Poznań Blocks | 27.20 | 27.19 | 27.13 | 26.98 | 26.79 |
| Poznań Blocks2 | 28.12 | 28.11 | 28.09 | 28.06 | 28.03 |
| Poznań Fencing2 | 28.60 | 28.50 | 28.43 | 28.36 | 28.35 |
| Poznań Service2 | 25.51 | 25.45 | 25.37 | 25.19 | 25.04 |
| *Average*: | 28.84 | 28.80 | 28.73 | 28.67 | 28.59 |

TABLE VIII
THE BITRATE AND QUALITY OF ENCODED VIRTUAL VIEWS. VIRTUAL VIEWS WERE SYNTHESIZED USING DEPTH MAPS ESTIMATED FOR DIFFERENT NUMBERS OF P-TYPE DEPTH FRAMES BETWEEN I-TYPE DEPTH FRAMES

| Test sequence | QP | Number of P type depth frames between I-type depth frames | | | | | | | | | |
|---|---|---|---|---|---|---|---|---|---|---|---|
| | | 0 | | 4 | | 9 | | 24 | | 49 | |
| | | Bitrate [Mbps] | PSNR [dB] | Bitrate [Mbps] | PSNR [dB] | Bitrate [Mbps] | PSNR [dB] | Bitrate [Mbps] | PSNR [dB] | Bitrate [Mbps] | PSNR [dB] |
| Ballet | 22 | 4.6 | 41.8 | 3.9 | 41.8 | 3.7 | 41.8 | 3.6 | 41.8 | 3.6 | 41.8 |
| | 27 | 1.7 | 39.8 | 1.5 | 39.9 | 1.4 | 39.9 | 1.4 | 40.0 | 1.4 | 40.0 |
| | 32 | 0.7 | 37.8 | 0.6 | 37.9 | 0.6 | 37.9 | 0.6 | 37.9 | 0.6 | 38.0 |
| | 37 | 0.3 | 35.9 | 0.3 | 35.9 | 0.3 | 35.9 | 0.3 | 35.9 | 0.3 | 35.9 |
| Break-dancers | 22 | 8.7 | 40.3 | 8.4 | 40.4 | 8.4 | 40.4 | 8.1 | 40.5 | 8.0 | 40.5 |
| | 27 | 2.8 | 38.2 | 2.7 | 38.3 | 2.7 | 38.3 | 2.7 | 38.3 | 2.6 | 38.5 |
| | 32 | 1.1 | 36.4 | 1.1 | 36.5 | 1.1 | 36.5 | 1.1 | 36.5 | 1.1 | 36.7 |
| | 37 | 0.5 | 34.7 | 0.5 | 34.8 | 0.5 | 34.8 | 0.5 | 34.9 | 0.5 | 35.1 |
| BBB Butterfly | 22 | 5.5 | 45.4 | 5.2 | 45.4 | 5.1 | 45.4 | 5.1 | 45.4 | 5.1 | 45.4 |
| | 27 | 2.5 | 42.0 | 2.4 | 42.0 | 2.4 | 42.1 | 2.3 | 42.1 | 2.3 | 42.1 |
| | 32 | 1.2 | 39.1 | 1.1 | 39.1 | 1.1 | 39.1 | 1.1 | 39.1 | 1.1 | 39.1 |
| | 37 | 0.5 | 36.5 | 0.5 | 36.5 | 0.5 | 36.5 | 0.5 | 36.6 | 0.5 | 36.5 |
| BBB Rabbit | 22 | 14.8 | 40.4 | 10.2 | 40.8 | 7.6 | 41.2 | 5.5 | 41.5 | 4.7 | 42.0 |
| | 27 | 6.0 | 36.0 | 5.0 | 36.6 | 3.8 | 37.1 | 2.7 | 37.4 | 2.3 | 37.8 |
| | 32 | 2.9 | 32.7 | 2.1 | 33.2 | 1.6 | 33.6 | 1.1 | 33.9 | 1.0 | 34.2 |
| | 37 | 0.9 | 30.2 | 0.7 | 30.5 | 0.6 | 30.8 | 0.4 | 31.0 | 0.4 | 31.2 |
| Poznań Blocks | 22 | 25 | 41.5 | 19.2 | 41.6 | 18.2 | 41.7 | 17.8 | 41.7 | 18.2 | 41.8 |
| | 27 | 9.5 | 38.1 | 7.5 | 38.2 | 7.0 | 38.4 | 6.8 | 38.3 | 7.0 | 38.4 |
| | 32 | 3.5 | 35.5 | 2.8 | 35.6 | 2.6 | 35.7 | 2.5 | 35.7 | 2.5 | 35.7 |
| | 37 | 1.3 | 33.4 | 1.0 | 33.5 | 0.9 | 33.6 | 0.9 | 33.6 | 0.9 | 33.6 |
| Poznań Blocks2 | 22 | 26.5 | 40.1 | 22.2 | 40.1 | 20.8 | 40.2 | 19.9 | 40.2 | 19.6 | 40.3 |
| | 27 | 7.7 | 37.6 | 6.5 | 37.7 | 6.0 | 37.8 | 5.7 | 37.9 | 5.6 | 37.9 |
| | 32 | 2.4 | 35.7 | 2.1 | 35.8 | 1.9 | 35.9 | 1.8 | 35.9 | 1.8 | 36.0 |
| | 37 | 0.8 | 34.1 | 0.7 | 34.1 | 0.6 | 34.2 | 0.6 | 34.2 | 0.6 | 34.3 |
| Poznań Fencing2 | 22 | 32.3 | 40.5 | 30.9 | 40.8 | 30.1 | 40.8 | 29.8 | 40.8 | 29.5 | 40.8 |
| | 27 | 14.5 | 36.7 | 13.4 | 37.6 | 12.9 | 37.7 | 12.6 | 37.7 | 12.4 | 37.7 |
| | 32 | 5.9 | 33.4 | 5.3 | 34.7 | 5.1 | 34.8 | 4.9 | 34.8 | 4.7 | 34.9 |
| | 37 | 2.0 | 32.0 | 1.8 | 32.5 | 1.7 | 32.5 | 1.7 | 32.5 | 1.6 | 32.7 |
| Poznań Service2 | 22 | 46.2 | 39.7 | 37.0 | 39.7 | 34.7 | 39.8 | 35.6 | 39.8 | 36.1 | 39.8 |
| | 27 | 18.1 | 36.3 | 14.8 | 36.5 | 13.8 | 36.7 | 14.4 | 36.6 | 14.5 | 36.6 |
| | 32 | 6.7 | 33.6 | 5.7 | 33.8 | 5.4 | 33.9 | 5.7 | 33.8 | 5.7 | 33.7 |
| | 37 | 2.3 | 31.6 | 2.0 | 31.7 | 1.8 | 31.7 | 2.0 | 31.5 | 2.1 | 31.4 |

REFERENCES

[1] G. Lafruit, M. Domański, K. Wegner, T. Grajek, T. Senoh, J. Jung, P. Kovács, P. Goorts, L. Jorissen, A. Munteanu, B. Ceulemans, P. Carballeira, S. García and M. Tanimoto, "New visual coding exploration in MPEG: Super-MultiView and Free Navigation in Free viewpoint TV," in *2016 Proceedings of the Electronic Imaging Conference: Stereoscopic Displays and Application*, San Francisco, 2016, pp. 1–9.

[2] F. Zilly, C. Riechert, M. Muller, P. Eisert, T. Sikora and P. Kauff, "Real-time generation of multi-view video plus depth content using mixed narrow and wide baseline," *Journal of Visual Communication and Image Representation*, vol. 25, no. 4, pp. 632–648, 2014.

[3] L. Fang, Y. Xiang, N. M. Cheung and F. Wu, "Estimation of Virtual View Synthesis Distortion Toward Virtual View Position," *IEEE Transactions on Image Processing*, vol. 25, no. 5, pp. 1961–1976, May 2016.

[4] Y. S. Kang and Y. S. Ho, "High-quality multi-view depth generation using multiple color and depth cameras," in *2010 IEEE International Conference on Multimedia and Expo*, Suntec City, pp. 1405–1410.

[5] S. Xiang, L. Yu, Q. Liu and Z. Xiong, "A gradient-based approach for interference cancelation in systems with multiple Kinect cameras," in *2013 IEEE International Symposium on Circuits and Systems (ISCAS2013)*, Beijing, pp. 13–16.

[6] G. Zhang, J. Jia, T. T. Wong and H. Bao, "Consistent Depth Maps Recovery from a Video Sequence," *IEEE Transactions on Pattern Analysis and Machine Intelligence*, vol. 31, no. 6, pp. 974–988, June 2009.

[7] O. Stankiewicz, K. Wegner, M. Tanimoto and M. Domański, "Enhanced Depth Estimation Reference Software (DERS) for Free-viewpoint Television", ISO/IEC JTC1/SC29/WG11 Doc. MPEG M31518, Geneva, 2013.

[8] L. Jorissen, P. Goorts, G. Lafruit and P. Bekaert, "Multi-view wide baseline depth estimation robust to sparse input sampling," in *2016 3DTV-Conference: The True Vision - Capture, Transmission and Display of 3D Video (3DTV-CON)*, Hamburg, pp. 1–4.

[9] V. Kolmogorov and R. Zabin, "What energy functions can be minimized via graph cuts?," *IEEE Transactions on Pattern Analysis and Machine Intelligence*, vol. 26, no. 2, pp. 147–159, Feb. 2004.

[10] Y. Boykov, O. Veksler and R. Zabih, "Fast approximate energy minimization via graph cuts," *IEEE Transactions on Pattern Analysis and Machine Intelligence*, vol. 23, no. 11, pp. 1222–1239, Nov 2001.

[11] C. L. Zitnick, S. B. Kang, M. Uyttendaele, S. Winder and R. Szeliski, "High-quality video view interpolation using a layered representation," ACM Transactions on Graphics, vol. 23, pp. 600–608, Aug. 2004.

[12] P. Kovacs, "[FTV AHG] Big Buck Bunny light-field test sequences". ISO/IEC JTC1/SC29/WG11, Doc. MPEG M35721, Geneva, 2015.

[13] M. Domański, A. Dziembowski, M. Kurc, A. Łuczak, D. Mieloch, J. Siast, O. Stankiewicz and K. Wegner, "Poznań University of Technology test multiview video sequences acquired with circular camera arrangement – "Poznan Team" and "Poznan Blocks" sequences", ISO/IEC JTC1/SC29/WG11, Doc. MPEG M35846, Geneva, 2015.

[14] M. Domański, A. Dziembowski, A. Grzelka, D. Mieloch, O. Stankiewicz and K. Wegner, "Multiview test video sequences for free navigation exploration obtained using pairs of cameras", ISO/IEC JTC1/SC29/WG11, Doc. MPEG M38247, Geneva, 2016.

[15] O. Stankiewicz, K. Wegner, M. Tanimoto and M. Domański, "Enhanced view synthesis reference software (VSRS) for Free-viewpoint Television", ISO/IEC JTC 1/SC 29/WG 11, Doc. MPEG M31520, Geneva, 2013.

[16] P. Yadati and A. M. Namboodiri, "Multiscale two-view stereo using convolutional neural networks for unrectified images," in *2017 Fifteenth IAPR International Conference on Machine Vision Applications (MVA)*, Nagoya, pp. 346–349.

[17] J. M. Fácil, A. Concha, L. Montesano and J. Civera, "Single-View and Multi-View Depth Fusion," *IEEE Robotics and Automation Letters*, vol. 2, no. 4, pp. 1994–2001, Oct. 2017.

[18] B. Tippetts, D. Jye Lee, K. Lillywhite and J. Archibald, "Review of stereo vision algorithms and their suitability for resource-limited systems", *Journal of Real-Time Image Processing*, vol. 11, no. 1, pp. 5–25, Jan. 2013.

[19] L. Fang, N. M. Cheung, D. Tian, A. Vetro, H. Sun and O. C. Au, "An Analytical Model for Synthesis Distortion Estimation in 3D Video," *IEEE Transactions on Image Processing*, vol. 23, no. 1, pp. 185–199, Jan. 2014.

[20] R. Achanta and S. Süsstrunk, "Superpixels and Polygons Using Simple Non-iterative Clustering," in *2017 IEEE Conference on Computer Vision and Pattern Recognition (CVPR)*, Honolulu, HI, pp. 4895–4904.

[21] H. Zhu, Q. Wang and J. Yu, "Occlusion-Model Guided Antiocclusion Depth Estimation in Light Field," *IEEE Journal of Selected Topics in Signal Processing*, vol. 11, no. 7, pp. 965–978, Oct. 2017.

[22] K. Calagari, M. Elgharib, P. Didyk, A. Kaspar, W. Matusik and M. Hefeeda, "Data Driven 2-D-to-3-D Video Conversion for Soccer," *IEEE Transactions on Multimedia*, vol. 20, no. 3, pp. 605-619, March 2018.

[23] L. Li, S. Zhang, X. Yu and L. Zhang, "PMSC: PatchMatch-Based Superpixel Cut for Accurate Stereo Matching," *IEEE Transactions on Circuits and Systems for Video Technology*, vol. 28, no. 3, pp. 679-692, March 2018.

[24] N. Vretos and P. Daras, "Temporal and color consistent disparity estimation in stereo videos," in *2014 IEEE International Conference on Image Processing (ICIP)*, Paris, pp. 3798–3802.

[25] M. F. Tappen and W. T. Freeman, "Comparison of graph cuts with belief propagation for stereo, using identical MRF parameters," in *Proceedings Ninth IEEE International Conference on Computer Vision*, Nice, France, 2003, pp. 900–906.

[26] Hua Shi, Hong Zhu, Jing Wang, Shun-Yuan Yu and Zheng-Fang Fu, "Segment-based adaptive window and multi-feature fusion for stereo matching", *Journal of Algorithms & Computational Technology*, vol. 10, no. 1, 2016, pp.3–11.

[27] M. Sizintsev and R. P. Wildes, "Spatiotemporal Stereo and Scene Flow via Stequel Matching," *IEEE Transactions on Pattern Analysis and Machine Intelligence,* vol. 34, no. 6, pp. 1206–1219, June 2012.

[28] W. Chen, M. J. Zhang and Z. X. Xiong, "Fast semi-global stereo matching via extracting disparity candidates from region boundaries," *IET Computer Vision*, vol. 5, no. 2, pp. 143–150, March 2011.

[29] O. Stankiewicz, M. Domański, A. Dziembowski, A. Grzelka, D. Mieloch and J. Samelak, "A Free-viewpoint Television System for Horizontal Virtual Navigation," *IEEE Transactions on Multimedia*, vol. 20, no. 8, pp. 2182-2195, Aug. 2018.

[30] M. Domański, O. Stankiewicz, K. Wegner and T. Grajek, "Immersive visual media – MPEG-I: 360 video, virtual navigation and beyond", in *IEEE International Conference on Systems, Signals and Image Processing IWSSIP 2017*, Poznań, Poland, pp. 1–9.

[31] Y. Zhang et al., "Light-Field Depth Estimation via Epipolar Plane Image Analysis and Locally Linear Embedding," *IEEE Transactions on Circuits and Systems for Video Technology*, vol. 27, no. 4, pp. 739–747, April 2017.

[32] D. Mieloch, A. Dziembowski, A. Grzelka, O. Stankiewicz and M. Domański, "Graph-based multiview depth estimation using segmentation," in *2017 IEEE International Conference on Multimedia and Expo (ICME),* Hong Kong, pp. 217–222.

[33] G. Nur, S. Dogan, H. K. Arachchi and A. M. Kondoz, "Impact of depth map spatial resolution on 3D video quality and depth perception," in *2010 3DTV-Conference: The True Vision - Capture, Transmission and Display of 3D Video*, Tampere, pp. 1–4.

[34] W. Liu, X. Chen, J. Yang and Q. Wu, "Robust Color Guided Depth Map Restoration," *IEEE Transactions on Image Processing*, vol. 26, no. 1, pp. 315–327, Jan. 2017.

[35] T. Emori, M. P. Tehrani, K. Takahashi, and T. Fujii, "Free-viewpoint video synthesis from mixed resolution multi-view images and low resolution depth maps," in *Proc. SPIE 9391, Stereoscopic Displays and Applications XXVI*, 2015, pp. 93911C.

[36] M. Camplani, T. Mantecón and L. Salgado, "Depth-Color Fusion Strategy for 3-D Scene Modeling With Kinect," *IEEE Transactions on Cybernetics*, vol. 43, no. 6, pp. 1560–1571, Dec. 2013.

[37] J. Hernández-Aceituno, R. Arnay, J. Toledo and L. Acosta, "Using Kinect on an Autonomous Vehicle for Outdoors Obstacle Detection," *IEEE Sensors Journal*, vol. 16, no. 10, pp. 3603–3610, 2016.

[38] A. Dziembowski, A. Grzelka, D. Mieloch, O. Stankiewicz, K. Wegner and M. Domański, "Multiview synthesis — Improved view synthesis for virtual navigation," in *2016 Picture Coding Symposium (PCS)*, Nuremberg, pp. 1–5.

[39] K. Muller, P. Merkle and T. Wiegand, "3-D Video Representation Using Depth Maps," *Proceedings of the IEEE*, vol. 99, no. 4, pp. 643–656, April 2011.

[40] M. Tanimoto, "FTV standardization in MPEG," in *2014 3DTV-Conference: The True Vision - Capture, Transmission and Display of 3D Video (3DTV-CON)*, Budapest, pp. 1–4.

[41] M. Tanimoto, M. Panahpour Tehrani, T. Fujii and T. Yendo, "FTV for 3-D Spatial Communication," *Proceedings of the IEEE*, vol. 100, no. 4, pp. 905–917, April 2012.

[42] C. Lee, A. Tabatabai, K. Tashiro, "Free viewpoint video (FVV) survey and future research direction," *APSIPA Transactions on Signal and Information Processing,* vol. 4, Oct. 2015.

[43] M. Domański, A. Dziembowski, D. Mieloch, A. Łuczak, O. Stankiewicz and K. Wegner, "A practical approach to acquisition and processing of free viewpoint video," in *2015 Picture Coding Symposium (PCS)*, Cairns, QLD, pp. 10–14.

[44] V. Kolmogorov and R. Zabih, "Multi-camera Scene Reconstruction via Graph Cuts," in *Proceedings of the 7th European Conference on Computer Vision-Part III (ECCV '02)*, London, UK, pp. 82–96.

[45] L. Fang, Y. Xiang, N. M. Cheung and F. Wu, "Estimation of Virtual View Synthesis Distortion Toward Virtual View Position," *IEEE Transactions on Image Processing*, vol. 25, no. 5, pp. 1961–1976, May 2016.

[46] O. Stankiewicz, M. Domański and K. Wegner, "Estimation of temporally-consistent depth maps from video with reduced noise," in *2015 3DTV-Conference: The True Vision - Capture, Transmission and Display of 3D Video (3DTV-CON)*, Lisbon, pp. 1–4.

[47] HEVC reference codec. [Online]. Available: https://hevc.hhi.fraunhofer.de/svn/svn_HEVCSoftware/

[48] G. Bjøntegaard, "Calculation of average PSNR differences between RD986 curves," ISO/IEC JTC 1/SC 29/WG 11, Doc. MPEG M15378, Austin, TX, 2001.

[49] S. Li, C. Zhu and M. T. Sun, "Hole Filling with Multiple Reference Views in DIBR View Synthesis," *IEEE Transactions on Multimedia*, vol. PP, no. 99, pp. 1-1.

[50] Z. Lee and T. Q. Nguyen, "Multi-Array Camera Disparity Enhancement," *IEEE Transactions on Multimedia*, vol. 16, no. 8, pp. 2168-2177, Dec. 2014.

[51] X. Suau, J. Ruiz-Hidalgo and J. R. Casas, "Real-Time Head and Hand Tracking Based on 2.5D Data," *IEEE Transactions on Multimedia*, vol. 14, no. 3, pp. 575-585, June 2012.

[52] D. Scharstein, H. Hirschmüller, Y. Kitajima, G. Krathwohl, N. Nesic, X. Wang, and P. Westling, "High-resolution stereo datasets with subpixel-accurate ground truth," in *German Conference on Pattern Recognition (GCPR 2014)*, Münster, Germany, pp.31-42.

[53] T. Schöps, J. L. Schönberger, S. Galliani, T. Sattler, K. Schindler, M. Pollefeys, A. Geiger, "A Multi-View Stereo Benchmark with High-Resolution Images and Multi-Camera Videos," in *2017 Conference on Computer Vision and Pattern Recognition (CVPR)*, Honolulu, Hawaii, pp. 2538-2547.

[54] T. Senoh, N. Tetsutani and H. Yasuda, "Depth Estimation and View Synthesis for Immersive Media," in *2018 International Conference on 3D Immersion (IC3D)*, Brussels, Belgium, 2018, pp. 1-8.

[55] Y. Chang, S. Kim and Y. Ho, "Depth upsampling methods for high resolution depth map," in *2018 International Conference on Electronics, Information, and Communication (ICEIC)*, Honolulu, HI, 2018, pp. 1-4.

[56] X. Jiang, M. L. Pendu and C. Guillemot, "Depth Estimation with Occlusion Handling from a Sparse Set of Light Field Views," in *2018 25th IEEE International Conference on Image Processing (ICIP)*, Athens, 2018, pp. 634-638.

[57] M. Domański et al., "Demonstration of a simple free viewpoint television system," in *2017 IEEE International Conference on Image Processing (ICIP)*, Beijing, 2017, pp. 4589-4591. doi: 10.1109/ICIP.2017.8297154

[58] O. Stankiewicz, G. Lafruit, M. Domański, "Multiview video: Acquisition, processing, compression and virtual view rendering," in *Image and Video Processing and Analysis and Computer Vision*, Eds.: R. Chellappa, S. Theodoridis, Academic Press Library in Signal Processing: vol. 6, Academic Press, 2018, pp. 3-74.

[59] X. Huang, J. Zhang, Q. Wu, L. Fan and C. Yuan, "A Coarse-to-Fine Algorithm for Matching and Registration in 3D Cross-Source Point Clouds," *IEEE Transactions on Circuits and Systems for Video Technology*, vol. 28, no. 10, pp. 2965-2977, Oct. 2018.

[60] Y. Pan, R. Liu, B. Guan, Q. Du and Z. Xiong, "Accurate Depth Extraction Method for Multiple Light-Coding-Based Depth Cameras," *IEEE Transactions on Multimedia*, vol. 19, no. 4, pp. 685-701, April 2017.

[61] "Overview of 3D video coding", ISO/IEC JTC1/SC29/WG11, Doc. MPEG N9784, Archamps, France, May 2008.

[62] J. Lei, L. Li, H. Yue, F. Wu, N. Ling and C. Hou, "Depth Map Super-Resolution Considering View Synthesis Quality," *IEEE Transactions on Image Processing*, vol. 26, no. 4, pp. 1732-1745, April 2017.

[63] G. Lee, B. Li and C. Chen, "Content-adaptive depth map enhancement based on motion distribution," in *2014 IEEE Visual Communications and Image Processing Conference*, Valletta, 2014, pp. 482-485.

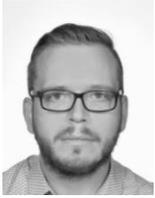
**Dawid Mieloch** received his M.Sc. and Ph.D. from Poznań University of Technology in 2014 and 2018, respectively. Currently, he is a research assistant at the Chair of Multimedia Telecommunications and Microelectronics. He is actively involved in ISO/IEC MPEG activities where he contributes to the development of the immersive media technologies. He has been involved in several projects focused on multiview and 3-D video processing. His professional interests include free viewpoint television, depth estimation and camera calibration.

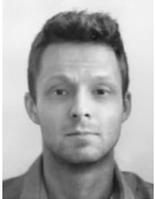
**Olgierd Stankiewicz** received his M. Sc. and Ph.D. from the Faculty of Electronics and Telecommunications, Poznań University of Technology in 2014. Currently, he is an assistant professor at the Chair of Multimedia Telecommunications and Microelectronics. In 2005 he won the second place in IEEE Computer Society International Design Competition (CSIDC), held in Washington D.C. He is actively involved in ISO standardization activities where he contributes to the development of the 3D video coding standards. In years 2011-2014 he was a coordinator of development of MPEG reference software for 3D-video coding standards based on AVC. Now he contributes to MPEG Free viewpoint TV and JPEG-PLENO standardization activities. He has published over ninety MPEG/JPEG standardization documents as well as about thirty papers on free view television, depth estimation, view synthesis and hardware implementation in FPGA. His professional interests include signal processing, video compression algorithms, computer graphics and hardware solutions.

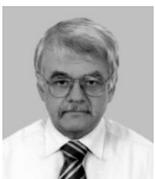
**Marek Domański** received M.Sc., Ph.D. and Habilitation degrees from Poznań University of Technology, Poland in 1978, 1983 and 1990, respectively. Since 1993, he is a professor at Poznań Univ. of Technology, where he leads Chair of Multimedia Telecommunications and Microelectronics. He coauthored one of the very first AVC decoders for tv set-top boxes (2004) as well as highly ranked technology proposals to MPEG for scalable video compression (2004) and 3D video coding (2011). He authored 3 books and over 300 papers in journals and conference proceedings. The contributions were mostly on image, video and audio compression, virtual navigation, free-viewpoint television, image processing, multimedia systems, 3D video and color image technology, digital filters and multidimensional signal processing. He was General Chairman/Co-Chairman and host of several international conferences: Picture Coding Symposium, PCS 2012; IEEE Int. Conf. Advanced & Signal based Surveillance, AVSS 2013, European Signal Processing Conference, EUSIPCO 2007; 73rd and 112nd Meetings of MPEG; Int. Workshop on Signals, Systems and Image Processing, IWSSIP 1997 and 2004; Int. Conf. Signals and Electronic Systems, ICSES 2004 and others. He served as a member of various steering, program and editorial committees of international journals and international conferences.